\documentclass[runningheads]{llncs}

\usepackage{iftex}

\usepackage{amsmath}
\usepackage{amsfonts}
\usepackage[inline]{enumitem}
\usepackage[T1]{fontenc}
\usepackage{graphicx}
\usepackage{color}
\usepackage{hyperref}
\usepackage{booktabs}

\usepackage{subcaption}
\usepackage{algorithm}
\usepackage{algorithmic}
\usepackage{empheq}

\DeclareMathOperator*{\argmax}{argmax}

\makeatletter
\newcommand{\myparagraph}[1]{\@ifclassloaded{nestpaper}{\paragraph{#1}}{\subsubsection{#1}}}
\makeatother

\begin{document}

\title{FORMICA: Decision-Focused Learning for Communication-Free Multi-Robot Task Allocation\thanks{DISTRIBUTION STATEMENT A. Approved for public release; distribution is unlimited. OPSEC10388}}
\titlerunning{FORMICA}

\author{Antonio Lopez \and
Jack Muirhead \and
Carlo Pinciroli\orcidID{0000-0002-2155-0445}}

\authorrunning{A. Lopez \emph{et al.}}

\institute{Dept.\ of Robotics Engineering, Worcester Polytechnic Institute, Worcester MA 01609, USA;\@ Email: \email{\{alopez3,jmuirhead,cpinciroli\}@wpi.edu}}

\maketitle
\begin{abstract}
Most multi-robot task allocation methods rely on communication to resolve conflicts and reach consistent assignments. In environments with limited bandwidth, degraded infrastructure, or adversarial interference, existing approaches degrade sharply. We introduce a learning-based framework that achieves high-quality task allocation without any robot-to-robot communication. The key idea is that robots coordinate implicitly by predicting teammates' bids: if each robot can anticipate competition for a task, it can adjust its choices accordingly. Our method predicts bid distributions to correct systematic errors in analytical mean-field approximations. While analytical predictions assume idealized conditions (uniform distributions, known bid functions), our learned approach adapts to task clustering and spatial heterogeneity. Inspired by \emph{Smart Predict-then-Optimize (SPO)}, we train predictors end-to-end to minimize \emph{Task Allocation Regret} rather than prediction error. To scale to large swarms, we develop a mean-field approximation where each robot predicts the distribution of competing bids rather than individual bids, reducing complexity from $O(NT)$ to $O(T)$. We call our approach \emph{FORMICA: Field-Oriented Regret-Minimizing Implicit Coordination Algorithm}. Experiments show FORMICA substantially outperforms a natural analytical baseline. In scenarios with 16 robots and 64 tasks, our approach improves system reward by 17\% and approaches the optimal MILP solution. When deployed on larger scenarios (256 robots, 4096 tasks), the same model improves performance by 7\%, demonstrating strong generalization. Training requires only 21 seconds on a laptop, enabling rapid adaptation to new environments.

\end{abstract}

\section{Introduction}
\label{sec:introduction}
Task allocation methods have achieved remarkable success in multi-robot systems, yet they share a common vulnerability: a strong dependency on communication. In distributed settings, methods such as the Consensus-Based Bundle Algorithm (CBBA)~\cite{choi2009consensus} and its variants~\cite{buckman2019partial,johnson2012asynchronous} require that robots eventually exchange winning bids and timestamps to resolve conflicts. When robots operate underwater where bandwidth is severely limited~\cite{berlinger2021implicit}, in disaster zones with damaged infrastructure, or under adversarial jamming, this dependency becomes a critical limitation. Reduced-communication approaches exist~\cite{dias2006market,quinton2023market}, yet to the best of our knowledge, no existing method achieves decentralized task allocation \emph{completely without communication} while retaining the performance benefits of optimization-based allocation.

The central insight of this paper is that robots can coordinate implicitly if they can predict each other's behavior. Consider two roommates dividing household chores without discussion: if each knows the other's preferences, they can coordinate without talking. This idea echoes classical notions of implicit coordination in biological swarms and recent work on machine theory of mind~\cite{albrecht2018autonomous,rabinowitz2018machine}. At swarm scale, however, manually engineering such predictive models becomes infeasible. Instead, robots must \emph{learn} to predict each other: A capability we call a \emph{swarm theory of mind}.

We present a market-based approach in which each robot computes bids for tasks and uses a learned model to estimate the competing bids of other robots. We call our method FORMICA: \emph{Field-Oriented Regret-Minimizing Implicit Coordination Algorithm}. Inspired by the \emph{Smart Predict-then-Optimize (SPO)} framework~\cite{elmachtoub2022smart,mandi2024decision}, FORMICA trains bid predictors to minimize \emph{Task Allocation Regret} (TAR), the loss of performance due to coordination failures. In task allocation, not all prediction errors are equal: underestimating a competitor's bid on a high-value task may cause conflicts and wasted resources; overestimating bids on low-value tasks may cause them to be ignored. By training predictors to minimize TAR directly, our approach learns to focus on the errors that matter.

\emph{Critically, all existing SPO and decision-focused learning methods address single-agent, centralized optimization problems}~\cite{el2019generalization,elmachtoub2020decision}. To the best of our knowledge, no prior work applies decision-focused learning to multi-agent coordination or any multi-robot system. To scale to large swarms, we employ a mean-field approximation: instead of predicting individual robot bids---which would require $O(NT)$ estimates per robot---each robot predicts the \emph{distribution} of bids across the swarm. This reduces complexity to $O(T)$ while preserving coordination.

Experiments show that our method substantially outperforms a natural analytical baseline in which robots select locally highest-valued tasks. In scenarios with 16 robots and 64 tasks, our approach improves system reward by approximately 17\% and approaches the optimal solution obtained via Mixed Integer Linear Programming. When deployed on much larger scenarios (256 robots, 4096 tasks), the same model improves performance by about 7\% over the analytical approach, demonstrating scalability and strong generalization.

\section{Related Work}
\label{sec:relatedwork}

\textbf{Multi-Robot Task Allocation.} Market-based methods like CBBA~\cite{choi2009consensus} achieve strong performance in decentralized settings but fundamentally require communication to resolve conflicts through bid exchanges and timestamps~\cite{buckman2019partial,johnson2012asynchronous}. Recent learning-based approaches combine GNNs with reinforcement learning~\cite{blumenkamp2024roboballet} or attention mechanisms~\cite{dai2025heterogeneous,goarin2024graph}, but all assume at least periodic communication or global broadcasts during execution. For communication-free allocation, Dai et al.~\cite{dai2019task} use game-theoretic equilibria, but cannot learn from data or optimize for allocation quality.

\textbf{Smart Predict-then-Optimize.} Elmachtoub and Grigas~\cite{elmachtoub2022smart} introduced SPO to address a fundamental mismatch: prediction models trained to minimize error often yield poor decisions when their outputs feed into optimization problems. SPO trains predictors to minimize \emph{decision regret} rather than prediction error. Extensions include generalization bounds~\cite{el2019generalization}, interpretable models~\cite{elmachtoub2020decision}, and applications to routing, scheduling, and inventory management~\cite{mandi2024decision}. \emph{Critically, all existing SPO work addresses single-agent, centralized problems.} To our knowledge, no prior work applies decision-focused learning to multi-robot systems or any multi-agent coordination problem.

\textbf{Communication-Free Coordination.} Several approaches achieve multi-robot coordination without explicit messaging. Swarm intelligence methods use stigmergy~\cite{theraulaz1999brief} or reactive rules~\cite{berlinger2021implicit,werfel2014designing}. Wang and Schwager~\cite{wang2016multi} enable manipulation through force sensing. For multi-agent prediction, work spans opponent modeling~\cite{carmel1996model,albrecht2018autonomous} to machine theory of mind~\cite{rabinowitz2018machine}. However, learning to predict bid values specifically for communication-free task allocation—and training these predictors end-to-end to minimize allocation regret—remains unexplored.

\section{Methodology}
\label{sec:methodology}

\subsection{Problem Formulation}
\label{sec:problem_formulation}

We focus on a task allocation problem in which $N \in \mathbb{R}^+$ robots must perform $T \in \mathbb{R}^+$ tasks, $T > N$. Each robot $i$ calculates the utility of performing task $j$, expressed as a bid $b_{i,j}$. Each robot has a maximum capacity $C_i$ to perform tasks. We denote with $x_{i,j}$ the binary decision variable set to 1 when robot $i$ performs task $j$, and 0 otherwise. Expressing this problem as Mixed-Integer Linear Programming (MILP) is a standard exercise; to comply with page limitations, we omit this formulation. This formalization fits scenarios such as garbage collection, resource collection, search-and-rescue missions, and area coverage.

The key twist is that robots cannot communicate, so they are not aware of each other's bids. To address this, robots must estimate other robots' bids and use these estimates to locally solve a task allocation problem. We denote robot $k$'s estimate of robot $i$'s bid for task $j$ as $\hat{b}_{i,j}^k(\theta_k)$. The parameters $\theta_k$ must be tuned to produce an effective estimator. Once tuned, each robot $k$ locally solves:
\begin{align}
  x^{k,\star} = \argmax_{x^k} &\quad \sum_{j=1}^T x_j^k \cdot \max\left(0, b_{k,j} - \hat{b}_{-k,j}^k\right) \tag{LOCOBJ}\label{eq:locobj}\\
  \text{subject to} &\quad \sum_{j=1}^T b_{k,j} x_j^k \le C_k, \notag
\end{align}
where $\max(0, b_j^k - \hat{b}_{-k,j}^k)$ represents the value robot $k$ associates to task $j$ compared to estimated competing bids.

The core problem is how to best tune the estimator parameters $\theta_k$ to maximize cumulative reward. We use a supervised learning approach that minimizes a well-defined learning metric. The main insight of this paper is how such metric is defined. An intuitive approach would train estimators $\hat{b}_{i,j}^k(\theta_k)$ to minimize prediction error (e.g., MSE), then use these estimates for task assignment. However, this approach does not consider that not all estimation errors are equal. Underestimating a competitor's bid on a high-value task causes conflicts and wasted resources; overestimating bids on low-value tasks causes them to be ignored entirely; errors on tasks where a robot is not competitive have negligible impact. Our insight is that estimators should minimize disagreement on final assignments, not prediction error. We seek an end-to-end optimization procedure that directly relates estimator parameters $\theta_k$ to coordination quality.


\subsection{The 2-Robot Case}
\label{sec:tworobotcase}
We first focus on a 2-robot scenario to build intuition on our approach. In this
scenario, the goal of robot $k$'s estimator is to output the best possible value
for $\hat{b}_{-k,j}^k(\theta_k)$, i.e., the bid estimate for the other robot. Instead
of minimizing the prediction error $\| b_{k,j}-\hat{b}_{-k,j}^k \|$, we aim to
minimize the \emph{Task Allocation Regret (TAR)}, i.e., the loss of performance
due to the tasks with no allocated robots. In practice, to minimize the TAR, we
will use a gradient-descent approach. This procedure will require calculating
the TAR's gradient with respect to the estimator parameters $\theta_k$. However,
working with discrete binary allocation values ($x_{k,j}$) does not yield
differentiable expressions. To circumvent this issue, we relax the optimization
problem.

\myparagraph{Problem Relaxation.}
Through Lagrangian relaxation, we reformulate the objective function in \eqref{eq:locobj} as follows:
\begin{equation*}
  \mathcal{L}_\text{REL} = \sum_{j=1}^T x_j^k \cdot \max \left( 0, b_{k,j} - \hat{b}_{-k,j}^k \right) - \lambda^k \left( \sum_{j=1}^T b_{k,j} x_j^k - C_k\right).
\end{equation*}
By taking the derivative of this expression with respect to $x_j^k$ and setting
it to 0, we can derive a condition for robot $k$ to take task $j$:
\begin{equation*}
  \frac{\partial \mathcal{L}_\text{REL}}{\partial x_j^k} = \left(b_{k,j}-\hat{b}_{-k,j}^k-\lambda^kb_{k,j}\right) = 0
  \qquad\Rightarrow\qquad
  b_{k,j}-\lambda^kb_{k,j} = \hat{b}_{-k,j}^k,
\end{equation*}
where $\lambda^kb_{k,j}$ represents the opportunity cost for robot $k$ to perform task
$j$. When $b_{k,j}-\lambda^kb_{k,j} > \hat{b}_{-k,j}^k$, robot $k$ should take task
$j$; in the opposite case, robot $k$ should ignore the task. Next, we consider a
differentiable, \emph{soft} allocation function $\tilde{x}_{k,j}(\theta_k)$ shaped as
a sigmoid in $[0,1] \subset \mathbb{R}$:
\begin{equation}
  \tag{SAF}\label{eq:softallocfunc}
  \tilde{x}_j^k(\theta_k) = 
  \frac{\exp\left\{\beta \cdot \left(b_{k,j}-\hat{b}_{-k,j}^k-\lambda^kb_{k,j}\right)\right\}}{\sum_{j'=1}^T \exp\left\{\beta \cdot \left(b_{k,j'}-\hat{b}_{-k,j'}^k-\lambda^kb_{k,j'}^k\right)\right\}} \cdot C_k.
\end{equation}
The quantity $\tilde{x}_j^k(\theta_k) / C_k$ can be interpreted as the probability
for robot $k$ to perform task $j$. With the normalization in \eqref{eq:softallocfunc},
the unweighted sum $\sum_j \tilde{x}_j^k(\theta_k)$ equals $C_k$. In our implementation
we instead apply a capacity-true normalization that rescales $\tilde{x}_j^k$ so that
the bid-weighted usage $\sum_j b_{k,j}\tilde{x}_j^k$ matches $C_k$, consistent with the
original knapsack constraint $\sum_{j=1}^T b_{k,j} x_{k,j} \le C_k$. Parameter
$\beta$ regulates the steepness of the sigmoid, with higher values making the
function look more step-like, and lower values producing slower ascents. Using
$\tilde{x}_j^k$, we define
\begin{equation}
  \tag{TAR-2}\label{eq:tar2}
  \mathcal{L}_\text{TAR}\left(\theta_1, \theta_2\right) = \sum_{j=1}^T b_j \cdot \left(1-\frac{\tilde{x}_j^1}{C_1}\right) \cdot \left(1-\frac{\tilde{x}_j^2}{C_2}\right).
\end{equation}

\myparagraph{Gradient Computation.}
To implement gradient descent and train the estimators, we calculate the
gradient of \eqref{eq:tar2}. Critically, the presence of the capacity
constraints makes task assignments interdependent, because each robot must
choose the best tasks within its capacity---even if a robot had the highest bid on
every task, it would not necessarily be able to perform everything. To cope with
this issue, our approach follows the primal-dual algorithm: The primal update
tunes $\theta_k$ (for fixed $\lambda^k$) and the dual update tunes $\lambda^k$ (for fixed
$\theta_k$). For the primal update, using the chain rule, we have
\begin{equation*}
  \frac{\partial\mathcal{L}_\text{TAR}}{\partial\theta_k} = \sum_{j=1}^T \frac{\partial\mathcal{L}}{\partial\tilde{x}_j^k} \cdot \frac{\partial\tilde{x}_j^k}{\partial\hat{b}_{-k,j}^k} \cdot \frac{\partial\hat{b}_{-k,j}^k}{\partial\theta_k}
                  = \beta \cdot \sum_{j=1}^T b_j \cdot \underbrace{\left( 1-\frac{\tilde{x}_j^{-k}}{C_{-k}} \right)}_{\hat{q}_j^k} \cdot \frac{\tilde{x}_j^k}{C_k} \cdot \left(1-\tilde{x}_j^k\right).
\end{equation*}
The term marked as $\hat{q}_j^k$ is the probability that the other robot does
not have a bid $\ge b_{k,j}$, as estimated by robot $k$. We highlight this term
because, in later parts of this manuscript, we will revisit its
definition. Conveniently, the magnitude of
$\partial\mathcal{L}_\text{TAR}/\partial\theta_k$ is high in those cases in which allocation decisions are
critical:
\begin{enumerate*}[label=\emph{\alph*}), itemjoin={{; }}]
\item when a task is valuable (large $b_j$)
\item when a task is at risk of not being taken by the other robot (low
  $\tilde{x}_j^{-k}$)
\item when robot $k$ is uncertain on whether a task is valuable (the expression
  $\tilde{x}_j^k \, (1-\tilde{x}_j^k)$ peaks for $\tilde{x}_j^k = 0.5$).
\end{enumerate*}
For the dual update, rather than calculating the full gradient, we follow the
classic approximation in optimization theory
$\mathcal{L}_\text{TAR}/\partial\lambda^k \approx \sum_{j=1}^T b_{k,j} \tilde{x}_j^k - C_k$
which ignores $\tilde{x}_j^k$'s dependency on $\lambda^k$. This approximation is
justified in practice for large enough values of $\beta$. It is also reasonable
because the main effect we want to achieve is updating $\lambda^k$ in the direction of
$\sum_{j=1}^T b_{k,j} \tilde{x}_j^k - C_k$. Thus, we update $\lambda^k$ with the rule
$\lambda^k \leftarrow \lambda^k + \alpha \sum_{j=1}^T b_{k,j}^k \tilde{x}_j^k - C_k$.

\subsection{The $N$-Robot Case}

\myparagraph{Naive Extension.}
\label{sec:Nrobotcase}
To extend our approach to the $N$-robot case, the theory presented so far allows for a naive strategy, in which a robot estimates the individual bids of all the other robots.
However, the computational complexity of the primal update is $O(N \cdot T)$,
which makes the approach amenable to small swarms only, also because it imposes
the estimator to calculate $N$ outputs for each task (e.g., we would train a
neural network with $N \cdot T$ outputs). This is not only a computational issue: As
$N$ increases, it becomes prohibitively harder for a robot to estimate other
robots' bids using locally available information. As the swarm size increases,
however, enabling coordination without communication becomes \emph{more}
desirable. In fact, for extremely large swarms, it would anyway be infeasible to
communicate, collect, and process all the necessary information in a timely
manner. This insight justifies searching for a more aggressive approximation.

\myparagraph{Mean Field Approximation.}
As $N \rightarrow \infty$, individual robot identity becomes irrelevant. Rather than estimating
individual robot bids, robot $k$ should estimate the \emph{empirical
  distribution} of bids across the swarm. In practice, for task $j$, we
introduce the \emph{bid density function} $\rho_j(b)$, which represents the
fraction of robots with bid $b$ for task $j$. In the mean field limit, we define
the highest bid on task $j$, $\hat{h}_j^k$, as an estimate of
$\sup\{b : \rho_j(b) > 0\}$. Using $\hat{h}_j^k$, we redefine
\eqref{eq:softallocfunc} as
\begin{equation}
  \tag{SAF-MF}\label{eq:softallocfunc-mf}
  \tilde{x}_j^k(\theta_k) = 
  \frac{\exp\left\{\beta \cdot \left(b_{k,j}-\hat{h}_j^k-\lambda^kb_{k,j}\right)\right\}}{\sum_{j'=1}^T \exp\left\{\beta \cdot \left(b_{k,j'}-\hat{h}_{j'}^k-\lambda^kb_{k,j'}^k\right)\right\}} \cdot C_k.
\end{equation}
To redefine the TAR gradient, we also need a way to characterize the probability
$\hat{q}_j^k$ that robot $k$ has the highest bid on task $j$. We start by
observing that, using $\rho_j(b)$, the fraction of robots with a bid higher than
$b_{k,j}$ is given by $p = \int_{b_{k,j}}^\infty \rho_j(b) db.$
If there are $N-1$ robots (excluding $k$), and each independently has
probability $p$ of having a bid $\ge b_{k,j}$, then $\hat{q}_j^k =
(1-p)^{N-1}$. For large $N$ and small $p$, $(1-p)^{N-1} \approx \exp\{-Np\}$, and
therefore: $\hat{q}_j^k = \exp \left\{ -N \int_{b_{k,j}}^\infty \rho_j(b) db \right\}.$
Using $\hat{q}_j^k$, we can redefine the TAR as
\begin{equation}
  \tag{TAR-MF}\label{eq:tarmf}
  \mathcal{L}_\text{TAR}(\theta_k) = \sum_{j=1}^T b_j \cdot \left( 1-\frac{\tilde{x}_j^k}{C_k} \right) \cdot \hat{q}_j^k,
\end{equation}
where both $\tilde{x}_j^k$ and $\hat{q}_j^k$ depend on $\theta_k$. The TAR gradient is therefore the sum of two contributions:
\begin{equation*}
  \frac{\partial\mathcal{L}_\text{TAR}}{\partial\theta_k} =
  \underbrace{\beta \sum_{j=1}^T b_j \cdot \hat{q}_j^k \cdot \left( 1-\frac{\tilde{x}_j^k}{C_k} \right) \cdot \frac{\tilde{x}_j^k}{C_k} \cdot \frac{\partial\hat{h}_j^k}{\partial\theta_k}}_\text{Term A}
  +
  \underbrace{\sum_{j=1}^T b_j \cdot \left( 1-\frac{\tilde{x}_j^k}{C_k} \right) \cdot \frac{\hat{q}_j^k}{\theta_k}}_\text{Term B}.
\end{equation*}
Term A has the same behavior as the gradient of \eqref{eq:tar2}, in that it
encourages accurate predictions of the maximum competing bid. Term B, in
contrast, is large when the task is valuable (large $b_j$) and robot $k$ is not
taking it ($1-\tilde{x}_j^k/C_k$ is large); thus, it encourages coverage
predictions on tasks robot $k$ is unwilling to take. The key advantage of this
approximation is that its computational complexity is only $O(T)$---robot $k$ has
only two quantities to calculate for each task: $\hat{h}_j^k$ and
$\hat{q}_j^k$. Under suitable regularity conditions, the mean field
approximation is appropriate, in that it is possible to train estimators for
$\rho_j(b)$, and therefore $\hat{h}_j^k$ and $\hat{q}_j^k$, that use only local
information. An example to illustrate the feasibility of this approximation is
presented next.

\myparagraph{Example: Garbage Collection.}
Consider a garbage collection scenario with uniform distribution of tasks and
robots. For a task at position $\mathbf{p}_j$, assume
$b_{i,j} \approx R_j / ( \| \mathbf{p}_i - \mathbf{p}_j \| + \epsilon)$, with
$R_j$ denoting the reward for completing task $j$ (e.g., how much garbage is
located at $\mathbf{p}_j$). The term $\epsilon$ is a small regularization constant that
prevents division by zero when calculating the bid. Alternatively, we can
understand a bid as the ratio between reward and cost. Then, $\epsilon$ can be
interpreted as the minimum cost to perform a task (time to load the garbage,
equipment setup, etc.). In this example, we assume task information to be
globally known, so each robot knows $R_j$ and $\mathbf{p}_j$
$\forall j \in [1,T]$. The definition of $b_{i,j}$ is purely geometric: the set of
positions with bid $b_{i,j}$ forms a circle (in 2D) or a sphere (in 3D) around
$\mathbf{p}_j$ with radius $r(b_{i,j}) = R_j / (b_{i,j} - \epsilon)$. Because of this
regularity, it becomes possible for a robot to estimate $\rho_j(b)$ using only
locally available information, such as the robot's current position and globally
known information on the tasks. In fact, we can derive an analytic expression
for $\rho_j(b)$ in this case.
For a swarm uniformly distributed in a 2D space $\Omega$, $\rho_j(b)$ is the fraction of points that yield a bid $b$ (a circumference of radius $r(b)$ over the total area of $\Omega$), multiplied by the conversion factor between distance and bids, $dr / db$:
\begin{align}
  r(b) &= \frac{R_j}{b}-\epsilon \qquad \Rightarrow \qquad \frac{dr}{db} = -\frac{R_j}{b^2} \notag \\
  \rho_j(b) &= \frac{2 \pi r(b)}{|\Omega|} \cdot \left| \frac{dr}{db} \right|
           = \frac{2\pi R_j}{|\Omega| \cdot b^2} \cdot \left( \frac{R_j}{b} - \epsilon\right) \tag{AMF}\label{eq:amf}.
\end{align}
The absolute value in $\left| dr/db \right|$ is needed to keep $\rho_j(b)$
positive, due to the inverse relationship between distance and bids.


\subsection{Training Algorithm}

Our training algorithm targets the mean-field approximation formulation and consists of two distinct phases. Phase 1 bootstraps the estimators using supervised learning, while Phase 2 refines them through decision-focused optimization.

\textbf{Phase 1: Supervised Learning of $\rho_j(b')$.} Rather than directly training estimators for $\hat{h}_j$ and $\hat{q}_j^k$, we train a neural network to predict the full bid density $\rho_j(b')$ for each task, from which both quantities can be analytically derived. For each training scenario, we: (1) compute all bids $b_{i,j}$ and normalize them to $b'_{i,j}$ using the characteristic length $\ell = \sqrt{|\Omega|/N}$; (2) construct empirical histograms over $B$ logarithmically-spaced bins to form target densities $\rho_j(b')$; (3) train the network to minimize cross-entropy loss:
$\mathcal{L}_{\text{CE}} = -\frac{1}{T}\sum_{j=1}^T \sum_b \rho_j(b') \log \hat{\rho}_j(b').$

We use a Set Transformer architecture \cite{lee2019set} that takes the unordered set of tasks as input (encoded as $[x_j/W, y_j/H, R_j/\bar{R}, 1/\ell, \log N]$, where $W \times H$ is the environment size and $\bar{R}$ is the average reward) and outputs per-task bid densities $\hat{\rho}_j(b') = \text{softmax}(\text{logits}_j)$ over the $B$ bins.

\textbf{Phase 2: Decision-Focused Learning.} After Phase 1 initialization, we refine the network parameters using TAR as the training objective. For each step, we sample a scenario and robot $k$, then: (1) compute $\hat{h}_j$ as a soft quantile of $\hat{\rho}_j(b')$ with subtractive tempering $\Delta b'$; (2) compute coverage probabilities $\hat{q}_j^k = \exp(-(N-1) \int_{b>b_{k,j}} \hat{\rho}_j(b)db)$; (3) perform soft allocation and compute TAR loss $\mathcal{L}_{\text{TAR}} = \sum_j R_j(1 - \tilde{x}_j^k/C_k)\hat{q}_j^k$. The primal update backpropagates through $\hat{\rho}_j$ and $\hat{h}_j$ only (not through $\hat{q}_j^k$), while the dual update adjusts $\lambda \leftarrow \lambda + \eta(\sum_j b_{k,j}\tilde{x}_j^k - C_k)$. Algorithm~\ref{alg:training} summarizes the complete procedure.

\begin{algorithm}[t]
\caption{Mean-Field FORMICA Training}
\label{alg:training}
\begin{algorithmic}[1]
\REQUIRE Training scenarios $\{(W_t, Z_t)\}$; bin config: $B{=}64$, $b'{\in}[0.02, 64]$; Phase 1: $p_1{=}400$ steps, $\alpha{=}3{\times}10^{-3}$; Phase 2: $p_2{=}1200$ steps, $\beta{=}3.5$, $q_h{=}0.70$, $\Delta b'{=}1.6$, $\gamma{=}0.08$, $\eta{=}10^{-3}$
\ENSURE Trained network parameters $\theta$
\STATE \textbf{Phase 1: Supervised Learning on $\rho(b')$}
\FOR{step = 1 to $p_1$}
    \STATE Sample scenario $(W, Z)$ with $N$ robots, $T$ tasks
    \FOR{task $j = 1$ to $T$}
        \STATE Compute bids $\{b_{i,j}\}$, normalize to $\{b'_{i,j}\}$, build histogram $\rho_j(b')$
    \ENDFOR
    \STATE $\hat{\rho}_j(b') \leftarrow \text{Network}(Z, N, |\Omega|)$;  $\mathcal{L}_{\text{CE}} = -\frac{1}{T}\sum_j \sum_b \rho_j(b') \log \hat{\rho}_j(b')$
    \STATE Update $\theta \leftarrow \theta - \alpha\nabla_\theta\mathcal{L}_{\text{CE}}$
\ENDFOR
\STATE
\STATE \textbf{Phase 2: Decision-Focused Learning}
\STATE Initialize $\lambda \leftarrow 0$
\FOR{step = 1 to $p_2$}
    \STATE Sample scenario $(W, Z)$; select random robot $k$; compute bids $\{b_{k,j}\}$
    \STATE $\hat{\rho}_j(b') \leftarrow \text{Network}(Z, N, |\Omega|)$
    \STATE $\hat{h}_j \leftarrow \text{SoftQuantile}(\hat{\rho}_j, q_h) - \Delta b'$ for all $j$
    \STATE $\hat{q}_j^k \leftarrow \exp(-(N-1)\int_{b>b_{k,j}} \hat{\rho}_j(b)db)$ for all $j$
    \STATE $\tilde{x}^k \leftarrow \text{SoftKnapsack}(b_k, \hat{h}, \lambda, \beta, C_k)$
    \STATE $\mathcal{L}_{\text{TAR}} \leftarrow \sum_j R_j(1 - \tilde{x}_j^k/C_k)\hat{q}_j^k$
    \STATE Update $\theta \leftarrow \theta - \alpha\nabla_\theta\mathcal{L}_{\text{TAR}}$ \hfill $\triangleright$ gradient only through $\hat{h}_j$
    \STATE Update $\lambda \leftarrow \lambda + \eta(\sum_j b_{k,j}\tilde{x}_j^k - C_k)$
\ENDFOR
\end{algorithmic}
\end{algorithm}

\section{Experimental Evaluation}

\begin{figure}[tb]
  \centering
  \subcaptionbox{FORMICA vs AMF vs MILP in the small scenario.\label{fig:ratio_hist_training}}{\includegraphics[height=3.2cm]{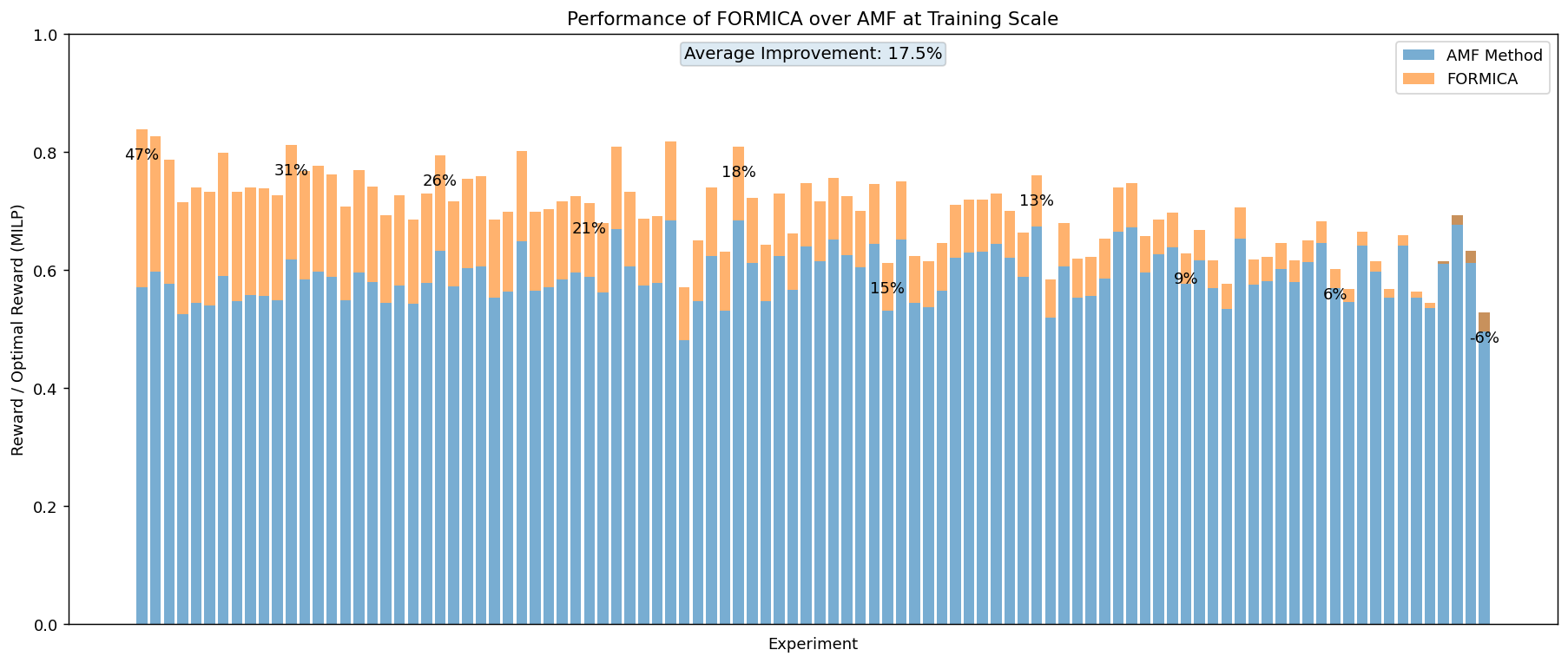}}
  \hfill
  \subcaptionbox{FORMICA vs AMF in the large scenario.\label{fig:ratio_hist_large}}{\includegraphics[height=3.2cm]{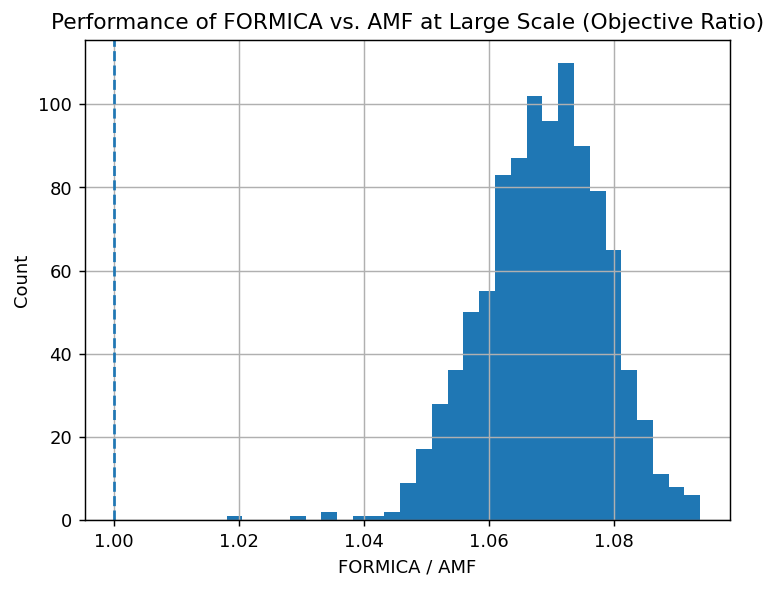}}
  \caption{Performance comparisons in the two scenarios.}
\end{figure}

\begin{table}[t]
\centering
\caption{Performance at training scale (16 robots, 64 tasks)}
\label{tab:training_scale}
\begin{tabular}{lcccc}
\toprule
Method & Objective & Coverage (\%) & Ratio to AMF & Ratio to MILP \\
\midrule
MILP           & $7.22 \pm 0.25$ & $89.6 \pm 4.1$ & $1.70 \pm 0.12$ & $1.00$         \\
FORMICA (ours) & $5.01 \pm 0.48$ & $54.3 \pm 5.6$ & $1.17 \pm 0.11$ & $0.69 \pm 0.07$ \\
AMF         & $4.27 \pm 0.28$ & $73.6 \pm 5.7$ & $1.00$          & $0.59 \pm 0.04$ \\
\bottomrule
\end{tabular}
\end{table}

\subsection{Experimental Setup}

We evaluate in a planar multi-robot task allocation scenario inspired by garbage collection. Each task $j$ at position $\mathbf{p}_j$ has reward $R_j \sim \mathcal{U}[6, 24]$. For robot $i$ at position $\mathbf{p}_i$, the bid is $b_{i,j} = R_j/(||\mathbf{p}_i - \mathbf{p}_j|| + \varepsilon)$ where $\varepsilon = 0.5$. Each robot has capacity $C_k = 0.5$ on the sum of accepted bids. The global objective is to maximize $\sum_j \max_i(b_{i,j} x_{i,j})$ subject to capacity constraints and at-most-one-robot-per-task.

\textbf{Training Configuration.} We train on $N = 16$ robots and $T = 64$ tasks in a $300 \times 200$ workspace with clustered task distribution (6 Gaussian clusters, $\sigma = 0.15 \cdot \min(W, H)$). See Algorithm~\ref{alg:training} for hyperparameters. The entire training took approximately 21 seconds on a laptop equipped with an 11th Gen Intel(R) Core(TM) i7-1185G7 @ 3.00GHz (3.00 GHz). This efficiency enables practical advantages:
\begin{enumerate*}[label=(\arabic*)]
\item rapid pre-deployment adaptation to specific environments;
\item potential online retraining during operations;
\item edge deployment without cloud infrastructure.
\end{enumerate*}
Communication-based methods like CBBA require no training, but demand continuous bandwidth. FORMICA eliminates such demand.

\textbf{Baselines.}
\begin{enumerate*}[label=(\arabic*)]
\item \textbf{Analytical Mean Field (AMF):} Each robot uses the closed-form bid density \eqref{eq:amf}. This represents the best performance achievable using perfect analytical predictions under idealized assumptions. Comparing against \eqref{eq:amf} isolates the value of learned corrections to handle non-uniform distributions, task clustering, and other deviations from mean-field assumptions.
\item \textbf{MILP}: For small-scale scenarios only, we solve a Mixed-Integer Linear Program using CBC to provide a near-optimal upper bound.
\end{enumerate*}

\textbf{Metrics.} We evaluate:
\begin{enumerate*}[label=(\arabic*)]
\item Global objective $\sum_j \max_i(b_{i,j} x_{i,j})$;
\item Task coverage (fraction with $\geq 1$ robot);
\item Performance ratio ($\text{obj}_{\text{method}} / \text{obj}_{\text{baseline}}$).
\end{enumerate*}


\subsection{Training Scale Results}

We evaluate on fresh scenarios from the training distribution: 16 robots, 64 tasks, $300 \times 200$ workspace.
Table~\ref{tab:training_scale} compares performance across $N = 100$ test scenarios. FORMICA achieves an average objective of $5.01 \pm 0.48$, compared to $4.27 \pm 0.28$ for the AMF baseline (approximately $17\%$ improvement; paired $t$-test, $p \ll 10^{-6}$) and $7.22 \pm 0.25$ for the MILP upper bound. This corresponds to $69.4 \pm 6.9\%$ of the MILP objective for FORMICA versus $59.2 \pm 4.3\%$ for AMF. FORMICA achieves 54\% coverage versus AMF's 74\%, yet delivers 17\% higher objective. This reveals decision-focused learning's key advantage: optimizing directly for reward rather than prediction accuracy leads to \emph{strategic task abandonment}. The method learns to sacrifice low-value coverage to secure high-value tasks: a clear instance of \emph{decision-focused learning}.

FORMICA outperforms AMF in $96/100 = 96\%$ of test scenarios (Fig.~\ref{fig:ratio_hist_training}). In clustered configurations, FORMICA successfully predicts high competition and reallocates robots to underserved regions, while the AMF baseline's analytical predictions fail to capture the actual bid distribution, leading to conflicts and missed coverage.

\subsection{Large-Scale Generalization}

We deploy the same network---trained only on 16-robot, 64-task scenarios---on $N = 256$ robots and $T = 4096$ tasks in a $3000 \times 2000$ workspace ($10\times$ larger in each dimension). This represents $16\times$ more robots, $64\times$ more tasks, and $100\times$ larger workspace area.

FORMICA outperforms AMF in every scenario we tested (1000/1000) with an average improvement of approximately ${\sim}6.9\%$ in the objective ($90.14 \pm 7.86$ vs.\ $84.32 \pm 7.09$; paired $t$-test, $p \ll 10^{-10}$; Fig.~\ref{fig:ratio_hist_large}, Table~\ref{tab:large_scale}). While the relative gain is smaller than at training scale---expected as the mean-field approximation becomes more accurate as $N$ increases---the consistency demonstrates successful transfer of learned corrections across scales. It also increases coverage from $39.2 \pm 3.5\%$ to $42.3 \pm 3.9\%$.

\begin{table}[t]
\centering
\caption{Large-scale generalization (256 robots, 4096 tasks)}
\label{tab:large_scale}
\begin{tabular}{lccc}
\toprule
Method & Objective & Coverage (\%) & Ratio to AMF \\
\midrule
FORMICA (ours) & $90.14 \pm 7.86$ & $42.3 \pm 3.9$ & $1.07 \pm 0.01$ \\
AMF            & $84.32 \pm 7.09$ & $39.2 \pm 3.5$ & $1.00$          \\
\bottomrule
\multicolumn{4}{l}{\footnotesize Note: MILP is computationally infeasible at this scale.}
\end{tabular}
\end{table}



\section{Conclusions}
\label{sec:conclusions}

We have presented FORMICA, a learning-based framework for multi-robot task allocation that achieves high-quality coordination without any robot-to-robot communication. Our approach addresses a critical vulnerability in existing methods: the strong dependency on stable, bandwidth-rich communication that cannot be guaranteed in underwater, disaster, or adversarial environments.

Our work makes three principal contributions. First, we introduce the first application of decision-focused learning to multi-agent coordination. By training predictors end-to-end to minimize Task Allocation Regret rather than prediction error, our method learns to focus on errors that matter most for achieving consensus. Second, we develop a mean-field approximation that reduces complexity from $O(NT)$ to $O(T)$, enabling a model trained on 16-robot scenarios to generalize to 256-robot swarms with consistent performance gains. Third, we demonstrate empirically that implicit coordination through learned behavioral prediction can rival communication-based methods, improving performance by approximately 17\% over an analytical baseline at training scale and 7\% when deployed on swarms $16\times$ larger. These contributions realize a form of \emph{swarm theory of mind}: the ability of each robot to internally model teammates' likely decisions without explicit information exchange.

Our work has several limitations that suggest future directions. We evaluate only homogeneous robot fleets with a single geometric bid model in static allocation scenarios. Extending to heterogeneous teams, dynamic task arrivals, alternative utility functions, and environments with obstacles or boundary effects would broaden applicability. Our implementation omits gradients through coverage probabilities for training stability, and exploring the full SPO gradient may improve performance in coverage-critical scenarios.

Several promising directions exist. Heterogeneous fleets would require ca\-pa\-bi\-li\-ty-conditioned bid distributions. Dynamic task allocation would enable adaptive search-and-rescue or persistent coverage applications. Alternative bid functions---energy-aware, deadline-sensitive, or multi-objective---would test generality. Perhaps most intriguing is the possibility of emergent specialization: decision-focused training might discover implicit role assignments that arise purely from coordination pressure. Beyond task allocation, our framework suggests a broader research agenda. The combination of mean-field approximation and decision-focused learning could apply to formation control, collaborative manipulation, or distributed sensing. 

\begin{credits}
\subsubsection{\ackname} This work was supported by the Automotive Research Center (ARC), a US Army Center of Excellence for modeling and simulation of ground vehicles, under Cooperative Agreement W56HZV-24-2-0001 with the US Army DEVCOM Ground Vehicle Systems Center (GVSC).

\subsubsection{\discintname}
The authors have no competing interests to declare that are relevant to the content of this article.
\end{credits}

\bibliographystyle{splncs04}
\bibliography{references}
\end{document}